\documentclass{article}
\usepackage[utf8]{inputenc}

\usepackage{graphicx, subfig}
\usepackage{amsmath}
\usepackage{amsthm}
\usepackage{amsfonts}
\usepackage{algorithm}
\usepackage[noend]{algpseudocode}
\usepackage{pbox}
\usepackage{geometry}
\usepackage{charter}
\usepackage{authblk}

\usepackage{hyperref}
\usepackage{fullpage}

\newcommand{\calM}{\mathcal{M}}
\newcommand{\calT}{\mathcal{T}}

\newcommand{\norm}[1]{\left\|#1\right\|}
\newcommand{\inner}[1]{\left<#1\right>}
\usepackage{xcolor}
\usepackage{comment}
\usepackage{verbatim}

\theoremstyle{definition}
\newtheorem{theorem}{Theorem}[section]
\newtheorem{lemma}[theorem]{Lemma}

\newtheorem{prop}[theorem]{Proposition}

\newtheorem{definition}{Definition}
\newtheorem{remark}{Remark}

\title{A Scale Invariant Flatness Measure for Deep Network Minima}
\author[1]{Akshay Rangamani \thanks{\url{rangamani.akshay@jhu.edu}}}
\author[2]{Nam H. Nguyen \thanks{\url{nnguyen@us.ibm.com}}}
\author[ ]{Abhishek Kumar \thanks{\url{abhishk@google.com}. Most of the work done while at IBM Research. Author is now at Google Brain.}}
\author[2]{Dzung Phan \thanks{\url{phandu@us.ibm.com}}}
\author[1,3]{Sang (Peter) Chin \thanks{\url{spchin@cs.bu.edu}}}
\author[1]{Trac D. Tran \thanks{\url{trac@jhu.edu}}}

\affil[1]{ECE Department\\
Johns Hopkins University}
\affil[2]{IBM Research}
\affil[3]{Department of Computer Science\\ Boston University}

\date{}

\begin{document}

\maketitle

\begin{abstract}
It has been empirically observed that the flatness of minima obtained from training deep networks seems to correlate with better generalization. However, for deep networks with positively homogeneous activations, most measures of sharpness/flatness are not invariant to rescaling of the network parameters, corresponding to the same function. This means that the measure of flatness/sharpness can be made as small or as large as possible through rescaling, rendering the quantitative measures meaningless. 
In this paper we show that for deep networks with positively homogenous activations, these rescalings constitute equivalence relations, and that these equivalence relations induce a quotient manifold structure in the parameter space. Using this manifold structure and an appropriate metric, we propose a Hessian-based measure for flatness that is invariant to rescaling. We use this new measure to confirm the proposition that Large-Batch SGD minima are indeed sharper than Small-Batch SGD minima.
\end{abstract}

\section{Introduction} \label{sec:intro}
In the past few years, deep learning \cite{lecun2015deep} has had empirical successes in several domains such as object detection and recognition \cite{krizhevsky2012imagenet, ren2015faster}, machine translation \cite{sutskever2014sequence,jean2014using}, and speech recognition \cite{hinton2012deep,sainath2013deep}, there is still a gap between theoretical bounds on the performance of deep networks and the performance of these networks in practice. Deep networks tend to be highly overparameterized, which means the hypothesis space is very large. 

However, optimization techniques such as stochastic gradient descent (SGD) are able to find solutions that generalize well, even if the number of training samples we have are far fewer than the number of parameters of the network we are training. This suggests that the solutions that we are able to retrieve certain desirable properties which are related to generalization.

Several empirical studies \cite{chaudhari2016entropy, keskar2016large} observe that the generalization ability of a deep network model is related to the spectrum of the Hessian matrix of the training loss at the solution obtained during training. It is also noted that solutions with smaller Hessian spectral norm tend to generalize better. These are popularly known as \emph{Flat Minima}, which have been studied since 1995 \cite{hochreiter1995simplifying, hochreiter1997flat}.

The flat minima heuristic is also related to a more formal framework for generalization -- PAC-Bayesian analysis of generalization behavior of deep networks. PAC-Bayes bounds \cite{dziugaite2017computing} are concerned with analyzing the behavior of solutions drawn from a posterior distribution rather than the particular solution obtained from empirical risk minimization, for instance. One posterior distribution the bounds are valid for are perturbations about the original solution obtained from empirical risk minimization. Neyshabur {\it et al.} relate the generalization of this distribution to the sharpness of the minima obtained \cite{neyshabur2017pac}. More recently, Wang {\it et al.} in \cite{wang2018identifying} provide formal connections between perturbation bounds and the Hessian of the empirical loss function and then propose a generalization metric that is related to the Hessian.

A number of quantitative definitions of flatness have been proposed both recently \cite{chaudhari2016entropy, keskar2016large} as well as in the early literature \cite{hochreiter1997flat}. These authors formalize the notions of ``flat" or ``wide" minima by either measuring the size of the connected region that is within $\epsilon$ of the value of the loss function at the minimum or by finding the difference between the maximum value of the loss function and the minimum value within an $\epsilon$-radius ball of the minimum. Note that the second notion of flatness is closely related to the spectral norm of the Hessian of the loss function at the minimum.
\medskip
\begin{definition}\label{def:sharpness}
If $B_2 (\epsilon, \theta)$ is the Euclidean ball of radius $\epsilon$ centered at a local minimum $\theta$ of a loss function $L$, then the $\epsilon$-sharpness of the minimum is defined as:
\[ \frac{\textrm{max}_{\theta^{'} \in B_2 (\epsilon, \theta)} L(\theta^{'}) - L(\theta)}{1+L(\theta)}. \]
\end{definition}
\medskip

By performing a second order Taylor expansion of $L$ around $\theta$, we can relate the $\epsilon$-sharpness to the spectral norm of the Hessian of $L$ at $\theta$ as follows
\[\frac{\textrm{max}_{\theta^{'} \in B_2 (\epsilon, \theta)} L(\theta^{'}) - L(\theta)}{1+L(\theta)} = \frac{||\nabla^2 L(\theta)||_2 \epsilon^2}{2(1+L(\theta))}. \]

However, \cite{dinh2017sharp}  show that deep networks with positively homogeneous layer activations (like the common ReLU activation, $\phi_{\textrm{rect}}(x) = \textrm{max}(0,x)$) can be rescaled to make their $\epsilon$-sharpness arbitrarily small or large with a simple transformation that implements the same neural network function but have widely different sharpness measures \cite{dinh2017sharp}.
To formalize this we consider a 2-layer neural network with parameters $\theta = (\theta_1, \theta_2)$ where the network is given by $y= \theta_2 \phi_{\textrm{rect}}(\theta_1 x)$. We can transform the parameters of the network by $\alpha > 0$ in the following manner: $T_{\alpha}(\theta) = (\alpha \theta_1, \alpha^{-1} \theta_2)$. We notice that for positively homogeneous activations, the networks parameterized by $\theta$ and $T_{\alpha}(\theta)$ implement the same function.
\medskip

\begin{theorem}
(Theorem 4 in \cite{dinh2017sharp}) For a one hidden layer rectified neural network of the form $y= \theta_2 \phi_{\textrm{rect}}(\theta_1 x)$ where $\theta = (\theta_1, \theta_2)$ is a minimum for $L$ such that $\nabla^2 L(\theta) \neq 0$, for any real number $M>0$, we can find a number $\alpha >0$ such that $||\nabla^2 L(T_{\alpha}(\theta))||_2 \geq M$. 
\end{theorem}
\medskip

This tells us that Hessian based measures like $\epsilon$-sharpness are not very meaningful since we can transform the parameters of the network to get as large or small a quantity as we want. This is also the case for other generalization metrics which are related to the Hessian, such as the one proposed in \cite{wang2018identifying}.

In this paper, we propose an alternative measure for quantifying the sharpness/flatness of minima of empirical loss functions. This measure is based on defining a quotient manifold of parameters which gives us a sharpness measure that is invariant to rescalings of the form described above. We use our sharpness measure to then test whether the minima obtained from large batch training are sharper than those obtained from small batch training.

The rest of the paper is organized as follows. In section \ref{sec:manifold}, we formalize the rescaling that can change the sharpness of minima without changing the function and show that the relation described by rescaling is indeed an equivalence relation, which in turn induces a manifold structure in the space of deep network parameters. In section \ref{sec:algorithm}, we describe an algorithm analogous to the power method that can be used to estimate the spectral norm of the Riemannian Hessian, which in turn can be employed as a measure of sharpness of the deep network minima. In section \ref{sec:lbsb}, we present several experimental results of applying our measure to small-batch vs large-batch training of various deep networks. Our results confirm that the geometric landscape of the loss function at small-batch minima are indeed flatter than that of large-batch minima.

\section{Characterizing a Quotient Manifold of Deep Network Parameters}\label{sec:manifold}
Let us define a neural network as a function $F: \mathbb{R}^{n_0} \rightarrow \mathbb{R}^{n_L}$ which takes an $n_0$-dimensional input and outputs an $n_L$-dimensional vector which could be a vector of class labels or a continuous measurement, depending on the task. We consider neural networks which consist of a series of nonlinear transformations, represented as
\[ F_{W}(\mathrm{x}) = W_L \phi_{L-1} ( W_{L-1} \phi_{L-2}(\ldots\phi_{1}(W_1 \mathrm{x}))). \]

Here $W_i \in \mathbb{R}^{n_i \times n_{i-1}}$ is a linear transformation, and $\phi_i$ is a positively homogeneous nonlinear function, usually applied pointwise to a vector. Each combination of a linear and nonlinear transformation is referred to as a ``layer", and the linear transformation $W_i$ is referred to as the parameter or weights of the layer. Even if $W_i$ has a matrix/convolutional structure, we will be concerned only with the vectorized version, $\textrm{vec} (W_i) \in \mathbb{R}^{d_i}$, which we will use interchangeably with $W_i$. First, we consider networks without bias vectors in each layer. We will extend our manifold construction to networks with bias at the end of this section.

Usually we have access to samples from a distribution $(\textrm{x},y) \sim \mathcal{D}$, and the way we train neural networks is by minimizing certain (convex) loss function that captures the distance between the network outputs and the target labels
\[ \textrm{min}_{W} \mathbb{E}_{(\textrm{x},y) \sim \mathcal{D}} \left[ \ell (F_W(\textrm{x}), y) \right]. \]

Due to the positive homogeneity we can scale the weights of the neural network appropriately to represent the same function with a different set of weights. This means there is a whole set of local minima that correspond to the same function, but are located at different points in the parameter space.
\medskip

\begin{prop} \label{prop:equivalence}
Let $W = (W_1, \ldots, W_L) \in \mathbb{R}^{d_1}_{*} \times \ldots \times \mathbb{R}^{d_L}_{*}$ be the parameters of a neural network with $L$ layers, and $\lambda = (\lambda_1, \ldots, \lambda_L) \in \mathbb{R}^{L}_{+}$ be a set of multipliers. Here $\mathbb{R}^{d_i}_{*} = \mathbb{R}^{d_i} \backslash \{0\} $. We can transform the layer weights by $\lambda$ in the following manner: $T_{\lambda}(W) = (\lambda_1 W_1, \ldots, \lambda_L W_L)$.
We introduce a relation from $\mathbb{R}^{n_1 \times n_0}_{*} \times \ldots \times \mathbb{R}^{n_L \times n_{L-1}}_{*}$ to itself, $W \sim Y$ if $\exists \lambda$ such that $Y = T_{\lambda} (W)$ and $\prod_{i=1}^L \lambda_i = 1$. Then, the relation $\sim$ is an equivalence relation.
\end{prop}

This equivalence relation is of interest to us because if $W \sim Y$, $F_{W}(\mathrm{x}) = F_{Y}(\mathrm{x})$ for all inputs $\mathrm{x} \in \mathbb{R}^{n_0}$. Denote $\overline{\mathcal{M}}_i = \mathbb{R}^{d_i}$ as the Euclidean vector space and the product manifold $\overline{\mathcal{M}} = \overline{\mathcal{M}}_1 \times \ldots \times \overline{\mathcal{M}}_L$ that covers the entire parameter space. We can use the equivalence relation defined in Proposition \ref{prop:equivalence} to obtain a quotient manifold induced by the equivalence relation $\mathcal{M} := \overline{\mathcal{M}}/\sim$.
\medskip

\begin{prop} \label{prop:manifold}
The set $\mathcal{M}:= \overline{\mathcal{M}}/\sim$ obtained by mapping all points within an equivalence class to a single point in the set has a quotient manifold structure, making $\calM$ a differentiable quotient manifold.
\end{prop}

Due to the space limitation, we leave the proof of Propositions \ref{prop:equivalence} and \ref{prop:manifold} in the Appendix.

Let $\mathcal{\pi}$ denote the mapping $\overline{\mathcal{M}} \rightarrow \mathcal{M}$ between the Euclidean parameter space and the quotient manifold. Given a point $W \in \overline{\mathcal{M}}$, $\mathcal{\pi}^{-1} (\mathcal{\pi} (W))$ is the equivalence class of $W$ and is also an embedded manifold of $\overline{\mathcal{M}}$. We have 
\begin{equation*}
\begin{split}
\mathcal{\pi}^{-1} (\mathcal{\pi} (W)) &= \bigg\{U: U_i = \lambda_i W_i; \lambda_i >0, \\
&\prod_{i=1}^L \lambda_i = 1, i=1,..., L \bigg\}.
\end{split}    
\end{equation*}

In order to impart a Riemannian structure to our quotient manifold, we need to define a metric on $\overline{\calM}$ that is invariant within an equivalence class.
\medskip

\begin{prop}\label{prop:metric}
Let $\eta_W$ and $\xi_W$ be two tangent vectors at a point $W \in \overline{\mathcal{M}}$. The Riemannian metric $\overline{g} : \mathcal{T}_{W}\overline{\mathcal{M}} \times \mathcal{T}_{W}\overline{\mathcal{M}}$ defined by:
\[ \overline{g}_{W}(\eta_W, \xi_W) = \sum_{i=1}^{L} \frac{\inner{\eta_{W_i}, \xi_{W_i} } }{||\textrm{vec}(W_i)||_2^2}
\]
is invariant within an equivalence class, and hence induces a metric for $\mathcal{M}$, $g_{\pi(W)} = \overline{g}_{W}$. Here $\langle \cdot, \cdot \rangle$ is the Euclidean inner product and $\eta_{W_i}, \; \xi_{W_i}$ are tangent vectors at a point $W_i \in \overline{\mathcal{M}}_i$.
\end{prop}
\begin{proof}
Let $U$ belong to the equivalent class $\mathcal{\pi}^{-1} (\mathcal{\pi} (W))$, and $\eta_W, \xi_W$ be tangent vectors in $\mathcal{T}_{W} \overline{\calM}$. Since $U$ and $W$ are in the same equivalence class, $\exists \lambda = (\lambda_1, \ldots, \lambda_L)$ such that $U = (U_1, ..., U_L) = (\lambda_1 W_1, ..., \lambda_L W_L)$ with $\prod_i \lambda_i = 1$. Using arguments similar to that presented in Example 3.5.4 in \cite{absil2009optimization}, the corresponding tangent vectors $\eta_U, \xi_U$ in $\mathcal{T}_{U} \overline{\calM}$ are related by the same scaling factor $\lambda$ between $U$ and $W$. 

Thus,
\[
\overline{g}_{U}(\eta_U, \xi_U) = \sum_{i=1}^{L} \frac{\inner{\lambda_i \eta_{W_i}, \lambda_i \xi_{W_i} } }{|| \lambda_i \textrm{vec}(W_i)||_2^2}
= \overline{g}_{W}(\eta_W, \xi_W),
\]
which completes the proof.
\end{proof}

One invariant property of the equivalence class is the product of the norms of all the layers. That is, if $U \in \mathcal{\pi}^{-1} (\mathcal{\pi} (W))$, then $\prod_i \norm{\textrm{vec}(U_i)}^2_2 = \prod_i \norm{\textrm{vec}(W_i)}^2_2$. For calculation convenience, we can replace the product by the sum by applying the log operator which gives $\sum_i \log \norm{\textrm{vec}(U_i)}^2_2 = \sum_i \log \norm{\textrm{vec}(W_i)}^2_2$.
\medskip

\begin{lemma} \label{lemma:vert}
The tangent space of $\mathcal{\pi}^{-1} (\mathcal{\pi} (W))$ at $U$ is $(\beta_1 U_1, ..., \beta_L U_L)$ with $\sum_i \beta_i = 0$.
\end{lemma}
\begin{proof}
Consider the curves $U_i(t) \in \overline{\calM}_i$ with $U_i(0) = U_i$, we have
\[
\sum_i \log \norm{\textrm{vec}(U_i (t))}^2_2 = \sum_i \log \norm{\textrm{vec}(W_i)}^2_2.
\]
Taking the derivative on both sides with respect to $t$ gives
\[
\sum_i \frac{\inner{\dot{U}_i(t), U_i(t)}}{\norm{\textrm{vec}(U_i (t))}^2_2} = 0.
\]
It is clear that $\dot{U}_i(t) = \beta_i U_i(t)$ with $\sum_i \beta_i = 0$ satisfies the above equation. Therefore the tangent space $\calT_U$ of $\mathcal{\pi}^{-1} (\mathcal{\pi} (W))$ contains all tangent vectors $\dot{U} = (\beta_1 U_1, ..., \beta_L U_L)$ with $\sum_i \beta_i = 0$.
\end{proof}

The tangent space to the embedded submanifold $\mathcal{\pi}^{-1} (\mathcal{\pi} (W))$ of $\overline{\mathcal{M}}$ is usually referred to as the Vertical Tangent space ($\mathcal{V}_W$) of the quotient manifold $\mathcal{M}$. The orthogonal complement of the vertical space from the tangent space $ \mathcal{T}_{W}\overline{\mathcal{M}} $ is referred to as the horizontal space $\mathcal{H}_W$. We note that all smooth curves $\gamma (t) : \mathbb{R} \rightarrow \overline{\calM}$ such that $\gamma(0) = W$ and $\dot{\gamma}(0) \in \mathcal{V}_W$, lie within the equivalence class $\mathcal{\pi}^{-1} (\mathcal{\pi} (W))$.

\subsection{Deep Networks with Biases}
A deep neural network with biases is a function $F: \mathbb{R}^{n_0} \rightarrow \mathbb{R}^{n_L}$ which takes an $n_0$-dimensional input and outputs an $n_L$-dimensional vector through a series of nonlinear transformations can be represented as
\begin{equation*}
\begin{split}
F_{(W,b)}(\mathrm{x}) = &W_L \phi_{L-1} ( W_{L-1} \phi_{L-2}(W_{L-2}\ldots\phi_{1}(W_1 \mathrm{x} + b_1) \\ &\ldots +b_{L-2})+b_{L-1} ) +b_L.
\end{split}
\end{equation*}

Here, $b_i \in \mathbb{R}^{n_i}$ are the bias parameters for each layer. Once again, due to the positive homogeneity of the nonlinear functions $\phi_i$, we can rescale the weights and biases of the network to obtain a different set of weights and biases that implement the same function.

Suppose we have $\lambda_i \in \mathbb{R}_{+}, i=1, \ldots, L$, such that $\prod_{i=1}^L \lambda_i = 1$. Consider the following transformation:
\begin{equation*}
\begin{split}
T_{\lambda} ((W,b)) &= (\lambda_L W_L, \ldots, \lambda_1 W_1, \\ &\prod_{i=1}^L \lambda_i b_L, \prod_{i=1}^{L-1} \lambda_i b_{L-1}, \ldots, \lambda_1 b_1).
\end{split}    
\end{equation*}

Now, if $(Y,c) = T_{\lambda}((W,b))$, then $F_{(W,b)} ( \mathrm{x}) = F_{(Y,c)} ( \mathrm{x})$ for all $ \mathrm{x}\in \mathbb{R}^{n_0}$. Let us denote $\overline{\calM}_i = \mathbb{R}^{d_i} \times \mathbb{R}^{n_i}$, as the Euclidean space for each layer. The product space $\overline{\calM} = \overline{\calM}_1 \times \ldots \times \overline{\calM}_L$ is the entire space of parameters for the neural networks with biases. Using arguments similar to Propositions \ref{prop:equivalence} and  \ref{prop:manifold}, we can see that this new transformation also introduces an equivalence relation on $\overline{\calM}$ and that $\calM := \overline{\calM}/\sim$ admits a quotient manifold structure. We modify Proposition \ref{prop:metric} slightly to get a new metric for the tangent space of $\overline{\calM}$.
\medskip

\begin{prop}\label{prop:metric_bias}
Since $\overline{\mathcal{M}}$ is a Euclidean space, its tangent space is also $\overline{\mathcal{M}}$. Let $\eta_{(W,b)}$ and $\xi_{(W,b)}$ be two tangent vectors at a point $(W,b) \in \overline{\mathcal{M}}$. The Riemannian metric $\overline{g} : \mathcal{T}_{{(W,b)}}\overline{\mathcal{M}} \times \mathcal{T}_{{(W,b)}}\overline{\mathcal{M}}$ defined by:
\[ \overline{g}_{{(W,b)}}(\eta_{(W,b)}, \xi_{(W,b)}) = \sum_{i=1}^{L} \left( \frac{\inner{\eta_{W_i}, \xi_{W_i} } }{||\textrm{vec}(W_i)||_2^2} + \frac{\inner{\eta_{b_i}, \xi_{b_i} } }{||b_i||_2^2} \right)
\]
is invariant within an equivalence class and hence induces a metric for $\mathcal{M}$, $g_{\pi(W,b)} = \overline{g}_{(W,b)}$. Here, $\langle \cdot, \cdot \rangle$ is the usual Euclidean inner product and $(\eta_{W_i}, \eta_{b_i})$ and $(\xi_{W_i}, \xi_{b_i})$ are tangent vectors at a point $(W_i,b_i) \in \overline{\mathcal{M}}_i$.
\end{prop}
\medskip

Let us now introduce a new invariant property of the equivalence class for network parameters with biases. First, for a point $(W,b)$ in the space of parameters, we know that $W = (W_1, \ldots, W_L)$ and $b= (b_1, \ldots, b_L)$. For each layer, let us define $\tilde{W_i} \in \mathbb{R}^{d_i + n_i}$ as follows 
\[
\tilde{W_i} =  \left\{
  \begin{array}{ll}
    [\textrm{vec}(W_i); \frac{b_i}{\norm{b_{i-1}}}], & \hbox{if } i > 1,\\[3mm] \newline 
    [\textrm{vec}(W_i); b_i], & \hbox{if } i = 1.
  \end{array}
\right. \vspace{-1mm}
\]
We then have that for each $(U,c) \in \mathcal{\pi}^{-1} (\mathcal{\pi} ((W,b)))$ if $\prod_i \norm{\tilde{U_i}}^2_2 = \prod_i \norm{\tilde{W_i}}^2_2$, which is the invariant property of the equivalence class. We can also get a description of the tangent space of $\pi^{-1}(\pi((W,b)))$ from the following lemma.
\medskip

\begin{lemma} \label{lemma:vert_bias}
The tangent space of $\mathcal{\pi}^{-1} (\mathcal{\pi} ((W,b)))$ at $(U,c)$ is $(\beta_1 U_1, ..., \beta_L U_L, \gamma_1 c_1, \ldots, \gamma_L c_L)$ with $\sum_i \beta_i = 0$, $\beta_i = \gamma_i - \gamma_{i-1}$.
\end{lemma}
\begin{proof}
Consider the curves $(U_i(t), c_i(t)) \in \overline{\calM}_i$ with $U_i(0) = U_i, c_i(0) = c_i$, we have
\[
\sum_i \log \norm{\tilde{U_i} (t)}^2_2 = \sum_i \log \norm{\tilde{W_i}}^2_2.
\]
\begin{equation*}
\begin{split}
\implies \sum_i \log \left( \norm{U_i (t)}^2_F + \frac{\norm{c_i(t)}^2}{\norm{c_{i-1}(t)}^2}\right) \\
= \sum_i \log \left(\norm{W_i}^2_F + \frac{\norm{b_i}^2}{\norm{b_{i-1}}^2} \right).
\end{split}
\end{equation*}

Taking the derivative on both sides with respect to $t$ gives
\begin{equation*}
\begin{split}
\sum_i \frac{1}{\norm{U_i (t)}^2_F + \frac{\norm{c_i(t)}^2}{\norm{c_{i-1}(t)}^2}}  \times \Bigg( \inner{\dot{U}_i(t), U_i(t)} \\
+ \frac{\inner{\dot{c}_i(t), c_i(t)}}{\norm{c_{i-1}(t)}^2} - \frac{\inner{\dot{c}_{i-1}(t), c_{i-1}(t)} \times \norm{c_{i}(t)}^2}{\norm{c_{i-1}(t)}^4} \Bigg) = 0.
\end{split}    
\end{equation*}

It is clear that $\dot{U}_i(t), \dot{c}_i(t) = \beta_i U_i(t), \gamma_i c_i(t)$ with $\sum_i \beta_i = 0$ and $\beta_i = \gamma_i - \gamma_{i-1}$ satisfies the above equation. Therefore the tangent space $\calT_{\mathcal{\pi}^{-1} (\mathcal{\pi} ((W,b)))} (U,c)$ at $U,c$ contains all tangent vectors $(\beta_1 U_1, ..., \beta_L U_L, \gamma_1 c_1, \ldots, \gamma_L c_L)$ with $\sum_i \beta_i = 0$ and $\beta_i = \gamma_i - \gamma_{i-1}$.
\end{proof}

\section{Measuring the Spectral Norm of the Riemannian Hessian} \label{sec:algorithm}
In the previous section, we introduced a quotient manifold structure that captures the rescaling that is natural to the space of parameters of neural networks with positively homogeneous activations. Now, similar to how the spectral norm of the Euclidean Hessian is used as a measure of sharpness, we can use the Taylor expansion of real-valued functions on a manifold to give us an analogous measure of sharpness using the spectral norm of the Riemannian Hessian. 

In this section, we will use normal symbols $W, f, \textrm{grad} f$ to denote points, functions, and gradients on the quotient manifold $\calM$, and overlines $\overline{W}, \overline{f}, \overline{\textrm{grad} f}$ to denote their lifted representations in the total manifold $\overline{\calM}$ (which is a vector space). If $\xi_W$ is a tangent vector in $\mathcal{T}_W \mathcal{M}$, then $\overline{\xi}_{\overline{W}}$ denotes the representation in $\mathcal{T}_{\overline{W}} \overline{\mathcal{M}}$ of the horizontal projection of $\xi_W$. The definition of Riemannian Hessian as per \cite{absil2009optimization} is as follows.
\medskip

\begin{definition}\label{def:hessvecprod}
For a real valued function $f$ on a Riemannian manifold $\mathcal{M}$, the \emph{Riemannian Hessian} is the linear mapping of $\mathcal{T}_x \mathcal{M}$ onto itself, defined by
\[ \textrm{Hess}f(W)[\xi_W] = \nabla_{\xi_W} \textrm{grad} f\]
for all $\xi_W \in \mathcal{T}_W \mathcal{M}$, where $\nabla$ is a Riemannian connection defined on $\mathcal{M}$.
\end{definition}
\medskip

To see how the Riemmannian Hessian is related to the flatness/sharpness of the function $f$ around a minimum $W$, we consider a retraction $R_W : \mathcal{T}_W \mathcal{M} \rightarrow \mathcal{M}$ which maps points in the tangent space to points on the manifold. For example, in a Euclidean space, $\mathcal{E}$, $R_x(\eta)=x+\eta$ is a retraction. The flatness/sharpness of a function around a minimum is defined (similar to Definition \ref{def:sharpness}) using the value of the function in a "neighborhood" of the minimum. To formalize what we mean by an $\epsilon$-neighborhood of $W$, it is the set of points that can be reached through a retraction using tangent vectors of norm at most $\epsilon$

\[ B_2 (\epsilon, W) = \{ R_W(\xi),  ||\xi||_g \leq \epsilon \} \]

Where $||\cdot||_g$ is the norm induced by the Riemannian metric $g$. This gives us the following flatness/sharpness measure:
\[ \frac{\textrm{max}_{W^{'} \in B_2 (\epsilon, W)} f(W^{'}) - f(W)}{1+f(W)}. \]

Using the fact that $\mathcal{T}_W \mathcal{M}$ is a vector space, and that $\hat{f}_W = f \circ R_W$ is a function on a vector space that admits a Taylor expansion, we get the following approximation for $f(W^{'})$ when $W^{'} \in B_2 (\epsilon, W)$, and $W^{'}=R_W(\xi_W)$:

\begin{equation*}
\begin{split}
f(W^{'}) \approx& f(W) + g( \textrm{grad}f(W), \xi_W) \\
& + \frac{1}{2} g(\xi_W, \textrm{Hess}f(W) [\xi_W])
\end{split}
\end{equation*}

Using the approximation, recognizing that at a minimum, $\textrm{grad} f(W)=0$, and using a Cauchy-Schwarz argument, we can bound the flatness/sharpness measure by the spectral norm of the Riemannian Hessian. We define it similar to the spectral norm of a linear map in Euclidean space.
\medskip

\begin{definition}
The spectral norm of the Riemannian Hessian of a function $f: \mathcal{M} \rightarrow \mathbb{R}$ is defined as
\[ ||\textrm{Hess}f(W)||_{2,g} = \underset{\xi_W \in \mathcal{T}_W \mathcal{M}, ||\xi_W||_g = 1}{\textrm{max}} g(\xi_W, \textrm{Hess}f(W) [\xi_W]) \]
\end{definition}
\medskip

With the definition of the spectral norm of the Riemannian Hessian, we now would like to be able to compute it for any function defined on a manifold. To achieve this, we present a Riemannian Power Method in Algorithm \ref{alg:riemannian_pm}.

\begin{algorithm}
\caption{Riemannian Power Method}\label{alg:riemannian_pm}
\begin{algorithmic}[1]
\Procedure{RiemannianPM}{$f,W$}
\State Initialize $\xi_{W}^0$ randomly in $\mathcal{T}_W \mathcal{M}$
\While{not converged} \;\; {(We use relative change in the eigenvector as a stopping criterion)}
\State $\xi_{W}^{t+1/2} \gets \textrm{Hess}f(W)[\xi_{W}^{t}]$
\State $\xi_{W}^{t+1}\gets \frac{\xi_{W}^{t+1/2}}{||\xi_{W}^{t+1/2}||_g}$
\State $t\gets t+1$
\EndWhile
\State \textbf{return} $\xi_{W}^{t}$ 
\EndProcedure
\end{algorithmic}
\end{algorithm}

\begin{remark}
Using Proposition 5.5.2 from \cite{absil2009optimization}, we have that:
\[ g(\xi, \textrm{Hess}f [\xi]) = \xi (\xi f) - (\nabla_{\xi} \xi ) f\]
Since we are only interested in computing the tangent vector $\xi_W$ in $\mathcal{T}_W \mathcal{M}$ that corresponds to the maximum eigenvalue of the linear map $\textrm{Hess} f(W)$ at the minimum, let us set $\xi$ to be a constant vector field, equal to $\xi_W$ at all points on the manifold.

We use a connection similar to the one defined in Theorem 3.4 of \cite{absil2004riemannian}, which means:
\begin{equation*}
\begin{split}
    \nabla_{\xi} \xi &= \mathcal{P}^\mathcal{H} \left( \overline{\nabla}_{\overline{\xi}_{\overline{W}}} \overline{\xi}_{\overline{W}}  \right) \\
    &= \mathcal{P}^\mathcal{H} \left( \frac{d}{dt}  \overline{\xi}_{\overline{W}+t\overline{\xi}_{\overline{W}}} \Bigg|_{t=0} \right) = 0
\end{split}
\end{equation*}

This means that $g(\xi, \textrm{Hess}f [\xi]) = \xi (\xi f)$. Now from the definition of the Riemmanian gradient of a function (equation 3.31 of \cite{absil2009optimization}), we have that:
\[ \xi f = g( \textrm{grad} f, \xi) \]
Considering $h = \xi f$ as another function on the manifold $\mathcal{M}$, we have:
\[ g(\xi, \textrm{Hess}f [\xi]) = \xi h =  g(\xi, \textrm{grad } g(\textrm{grad} f, \xi) )\]
Which means the Hessian vector product can be computed as:
\[ \textrm{Hess}f(W) [\xi_W] = \textrm{grad } g(\textrm{grad} f(W), \xi_W)\]
\end{remark}

\begin{remark}
While we have specified the definition of the spectral norm of the Riemannian Hessian  and the algorithm to compute it in general for any Riemannian manifold, we recall that we are dealing with a quotient manifold of neural network parameters. In order to implement our algorithms on a computer, we use the lifted representations of points and tangent vectors in the total manifold $\overline{\mathcal{M}}$. The lifted representation of a point $W$ is the parameter vector $\overline{W}$. The lifted representation of a tangent vector $\xi_W$ is the projection of the representation of a tangent vector $\tilde{\xi}_{\overline{W}} \in \mathcal{T}_{\overline{W}} \overline{\mathcal{M}}$ into the horizontal space $\mathcal{H}_W$, i.e., $\overline{\xi}_{\overline{W}} = \mathcal{P}^\mathcal{H}(\tilde{\xi}_{\overline{W}})$.

However, since the neural network loss functions that we would like to estimate the Hessian spectral norm for are constant within an equivalence class, their gradients are always zero along tangent directions within the vertical tangent space, which means the gradient lies in the horizontal space, and the projection of the lifted representation is unnecessary in practice. This is also true for the Hessian-vector product, which is computed as the gradient of the inner product between the gradient and the tangent vector.

For the sake of completion, the Riemannian gradient is computed as:
\[ \overline{\textrm{grad} f(W)} = \overline{G}^{-1}_{\overline{W}} \textrm{EGrad} \overline{f} (\overline{W}) \]

In the above equation, $\textrm{EGrad} \overline{f} (\overline{W})$ is the Euclidean gradient of the function $\overline{f}$, and  $\overline{G}^{-1}_{\overline{W}}$ is the inverse of the matrix representation of the metric at $\overline{W}$. The Euclidean gradient is easily computed using backpropagation. The inverse metric is given by $\overline{G}^{-1}_{\overline{W}} = \textrm{diag} (\ldots, ||\textrm{vec}(W_i)||^2 I_{d_i \times d_i}, \ldots )$.

\end{remark}

\subsection{Simulations}
To validify Algorithm \ref{alg:riemannian_pm}, we consider two deep network architectures described in Table \ref{tab:sim_archs}. For each architecture, we generate a synthetic dataset containing $N=500$ samples in $\mathbb{R}^{784}$ which belong to one of 10 different classes with randomly generated class labels. For each network, we consider softmax cross-entropy as the loss function. 

\begin{table}[h]
\begin{center}
\begin{tabular}{|c|c|} \hline
\textbf{Network} & \textbf{Architecture} \\ \hline
$F_1$     &  \pbox{60mm}{[FC($784, 300$), FC($300,100$), FC($100,10$)]}\\ \hline
$C_1$     & \pbox{60mm}{[conv($5,5,10$), conv($5,5,20$), FC($320, 120$), FC($120, 84$), FC($84, 10$)]}\\ \hline
\end{tabular}
\caption{Network Architectures for Simulations}
\label{tab:sim_archs}
\end{center}
\end{table}

We compute the spectral norms of the Hessians of their losses at different points within the equivalence class by considering $(Y,c) = T_{\lambda}((W,b))$ for different settings of $\lambda$. Let $\sigma_{(W,b)}$ be the spectral norm computed at $(W,b)$, and $\sigma_{(Y,c)}$ be the spectral norm computed at $(Y,c)$. We define the relative difference between the two measurements as follows:
\[\textrm{\texttt{Relative Difference}} = \frac{|\sigma_{(W,b)} - \sigma_{(Y,c)}|}{\sigma_{(W,b)}}\]
Results for $F_1$ are reported in Table \ref{tab:res_f1} whereas results for $C_1$ are reported in Table \ref{tab:res_c1}. 

\begin{table}[h]
\begin{center}
\begin{tabular}{|c|c|} \hline
$\lambda$ & Relative Difference \\ \hline
$(5, 4, \frac{1}{20})$ & $1.7 \times 10^{-7}$\\ \hline
$(100, 30, \frac{1}{3000})$ & $7.17 \times 10^{-7}$\\ \hline
\end{tabular}
\caption{\texttt{Relative Difference} in Spectral Norms for $F_1$ under different transformations}
\label{tab:res_f1}
\end{center}
\end{table}

\begin{table}[h]
\begin{center}
\begin{tabular}{|c|c|} \hline
$\lambda$ & Relative Difference \\ \hline
$(5, 4, 3, 2,\frac{1}{120})$ & $1.28 \times 10^{-7}$\\ \hline
$(50, 24, 30, \frac{1}{6}, \frac{1}{6000})$ & $5.1 \times 10^{-6}$\\ \hline
\end{tabular}
\caption{\texttt{Relative Difference} in Spectral Norms for $C_1$ under different transformations}
\label{tab:res_c1}
\end{center}
\end{table}

\begin{figure}[h]
    \centering
    \includegraphics[scale=0.7]{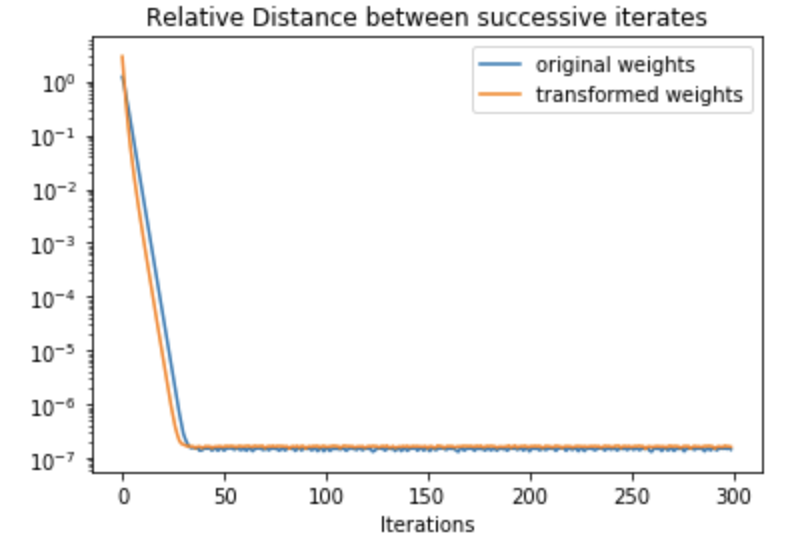}
    \caption{Convergence of Algorithm \ref{alg:riemannian_pm} for a synthetic dataset for an $F_1$ network}
    \label{fig:sim_f1}
\end{figure}

In Figure \ref{fig:sim_f1}, we can observe how our power method based algorithm converges for an $F_1$ network. From the tables, we notice that the spectral norm that we compute using the eigenvectors from Algorithm \ref{alg:riemannian_pm} is invariant to transformations within the equivalence class. That is, the values for \texttt{Relative Difference} are small. We can substitute in the spectral norm that we compute on the manifold into Definition \ref{def:sharpness} in order to come up with a measure of flatness that is invariant to rescaling.

\section{Large Batch vs Small Batch Training} \label{sec:lbsb}
One context in which the flatness of deep network minima has been suggested to correlate with better generalization is in large-batch training vs small-batch training of neural networks. In an empirical study, \cite{keskar2016large} observe that small-batch gradient methods with 32-512 samples per batch tend to converge to flatter minima than large-batch methods which have batch sizes of the order of $1000$s of samples. However, since \cite{dinh2017sharp} have shown that measures of flatness can be gamed by rescaling the network appropriately, we cannot trust the current quantitative measures to compare the sharpness of small-batch vs large-batch minima. Instead, we use the spectral norm of the Riemannian Hessian as a measure of sharpness and compare small-batch gradient based methods to large-batch methods.

\subsection{Datasets and Network Architectures}
Similar to \cite{keskar2016large}, we consider two datasets -- MNIST \cite{lecun1998mnist} and CIFAR-10 \cite{krizhevsky2009learning} -- with two different network architectures for each dataset. 

For MNIST, we used a fully connected deep network (MNIST-FC) with $5$ hidden layers of $512$ neurons each. In addition to this, we used a convolutional network based on the LeNet architecture \cite{lecun1998mnist}. This network has two convolutional-pooling layers, followed by two fully connected layers of $120$ and $84$ neurons before the final output layer with $10$ neurons.

For CIFAR-10, we considered a shallow convolutional network with an AlexNet-type architecture \cite{krizhevsky2012imagenet} and a deep convolutional network with a VGG16-type architecture \cite{simonyan2014very}.

In order to test our measure, we did not use layers which are not positively homogeneous like Local Response Normalization. Even though Batch Normalization layers are compatible with our manifold structure (if we consider the trained BN layer parameters as part of the network parameters), we did not use them in order to keep the experiments simple.

\subsection{Results}
Our goal in this set of experiments is not to achieve state of the art performance on these datasets. Instead, we are interested in characterizing and contrasting the solutions obtained using small-vs-large-batch gradient based methods. For each network architecture and dataset, we trained the network to $100\%$ training accuracy using SGD or Adam, resulting in training cross-entropy loss values in the order of $10^{-4}$\footnote{All code used to run the experiments can be found at \url{https://github.com/akshay-r/scale-invariant-flatness}.}. For MNIST we used batch sizes of $256$ and $5000$ samples for the small batch and large batch training respectively, while for CIFAR-10, we used batch sizes of $256$ and $2000$. The MNIST networks were trained using Adam while the CIFAR-10 networks were trained using SGD. The learning rate used for small batches was $0.0001$ while a learning rate of $0.001$ was used for large batches. In the case of both MNIST and CIFAR-10, we computed our flatness measure on the empirical loss on the training set at the minima obtained through the training process. Due to memory issues in the case of CIFAR-10, we limit ourselves to using $10000$ training examples instead of the entire training set for computing the flatness measure.

\begin{figure}[!t] 
\subfloat[AlexNet]{
\includegraphics[scale=0.35]{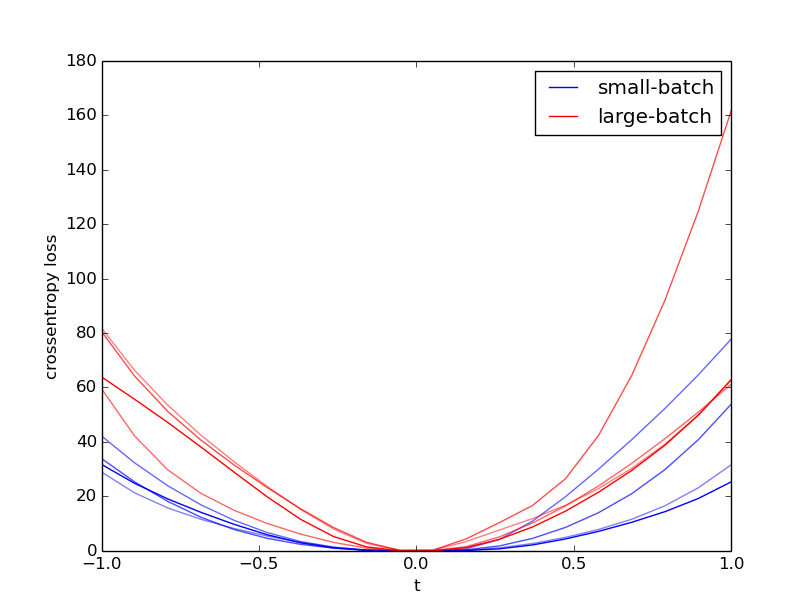}
\label{fig:alexnet}}
\vspace{5mm}
\subfloat[VGG16]{
\centering
\includegraphics[scale=0.35]{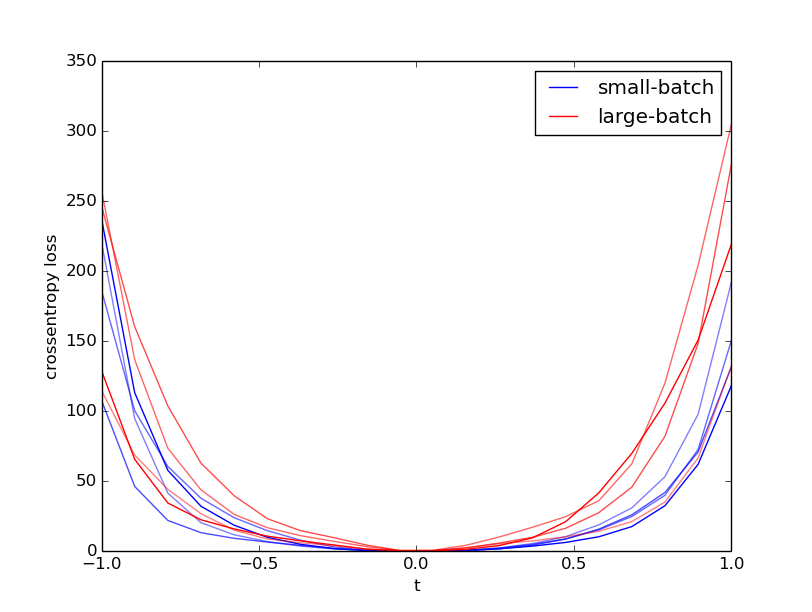}
\label{fig:vgg}}
\vspace*{\fill}
\caption{Parametric line plots for convolutional networks trained on CIFAR-10}
\label{fig:plots}
\end{figure}

Five different repetitions of these experiments were conducted, from different random initializations. We first generate parametric line plots along different random directions for AlexNet and VGG16. These plots are shown in Figure \ref{fig:plots}. These plots are \emph{layer normalized} \cite{li2017visualizing}, which means that the random directions chosen are scaled according to the norms of the layers of the trained networks. More precisely, if the minimum obtained from training AlexNet/VGG is $W = (W_1, \ldots, W_L)$, we generate random direction $V = (V_1, \ldots, V_L)$, and plot the loss along the curve $\hat{W}(t)$ for $t \in [-1,1]$. Here $\hat{W}(t)$ is given by:
\[ \hat{W}(t) = \left( \ldots, W_i + t \times \frac{||\textrm{vec}(W_i)||_2}{||\textrm{vec}(V_i)||_2} V_i, \ldots \right). \]

From the plots we see that the large-batch plots are above the small-batch plots, indicating that the large-batch minima are sharper than the small-batch counterparts.

Now, in order to quantify the sharpness and see how it correlates with generalization, we report the test accuracy and spectral norm of the Hessian at minima for each of the four networks trained on their respective datasets in Table \ref{tab:lb_vs_sb}. We observe that the estimated spectral norms for the large-batch minima are orders of magnitude larger than those of the small-batch minima for every network and dataset. This also correlates with test accuracy, with the small-batch minima having better generalization abilities.

We see that the difference in the spectral norm is 3-4 orders of magnitude. However, the same effect is not observed in the parametric line plots in Figure \ref{fig:plots}. This can be attributed to the fact that the spectral norm is only indicative of the sharpness along one particular direction or subspace in a very high dimensional parameter space. The parametric line plots are plotted along random directions, and thus we should not expect that the difference in sharpness will be of the same order of magnitude along all or even most random directions.

\begin{table}[h]
    \centering
    \begin{tabular}{|c|c|c|}\hline
    \textbf{Batch Size} & \textbf{Test Accuracy} & \textbf{Spectral Norm} \\ \hline
    \multicolumn{3}{|c|}{\textbf{MNIST / Fully-Connected}}  \\ \hline
    256 & $98.6 \pm 0.1 \%$ & $0.096 \pm 0.063$ \\ \hline 
    5000 & $96.8  \pm 0.2 \%$ & $73805 \pm 28984$ \\ \hline
    \multicolumn{3}{|c|}{\textbf{MNIST / LeNet}}  \\ \hline
    256 & $98.4 \pm 0.05 \%$ & $6e-5 \pm 4e-5$ \\ \hline 
    5000 & $97.8 \pm 0.1 \%$ & $4.03 \pm 1.95$ \\ \hline
    \multicolumn{3}{|c|}{\textbf{CIFAR-10 / AlexNet}} \\ \hline
    256 & $73.16 \pm 0.53 \%$ & $1790.34 \pm 221.47$ \\ \hline 
    2000 & $71.47 \pm 0.33 \%$ & $28533.22 \pm 2005.78$ \\ \hline
    \multicolumn{3}{|c|}{\textbf{CIFAR-10 / VGG16}} \\ \hline
    256 & $75.85 \pm 0.41 \%$ & $68.68 \pm 10.69$ \\ \hline 
    2000 & $68.55 \pm 0.1 \%$ & $25395.22 \pm 4025.52$ \\ \hline
    \end{tabular}
    \caption{Test Accuracy and Spectral Norm of Hessian at Minima for different trained networks. MNIST-FC and LeNet are trained and tested on MNIST, while AlexNet and VGG16 are trained and tested on CIFAR-10.}
    \label{tab:lb_vs_sb}
\end{table}

\section{Related Work}
In this paper, we have proposed a Hessian based measure for the sharpness of minima, which follows pioneering works in \cite{hochreiter1997flat} and \cite{keskar2016large} in attempting to measure the sharpness/flatness of deep network minima. As we noted in section \ref{sec:intro}, flatter minima are believed to be robust to perturbation of the neural network parameters. \cite{novak2018sensitivity} connect generalization to the sensitivty of the network to perturbations to the inputs. In a recent work, \cite{wang2018identifying} obtain a measure of generalization that is also related to the Hessian at the minima, but still have not resolved the rescaling issue that results in arbitrarily large or small Hessian spectra for the same neural network function.

Riemannian approaches to training neural networks have mostly focused on batch normalization \cite{cho2017riemannian, hoffer2018norm}. Since batch norm layers are invariant to scalings of the linear layers that precede them, a common approach is to restrict the weights of the linear layers to the manifold of weight matrices with unit norm, or an oblique manifold \cite{huang2017projection}, or the Stiefel manifold. To the best of our knowledge, we are the first to propose a quotient manifold of neural network parameters and successfully employ it to resolve the question of how to accurately measure the Hessian of the loss function at minima.

\section{Conclusion and Future Work}
In this paper, we observe that natural rescalings of neural networks with positively homogeneous activations induce an equivalence relation in the parameter space which in turn leads to a quotient manifold structure in the parameter space. We provide theoretical justification for these claims and then adopt the manifold structure to propose a Hessian based sharpness measure for deep network minima. We provide an algorithm to compute this measure and apply this technique to compare minima obtained using large-batch and small-batch gradient based methods.

We believe this quotient manifold view of the parameter space of neural networks can have implications for training deep networks as well. While balanced training procedures like weight normalization \cite{salimans2016weight} and Path-SGD \cite{neyshabur2015path} have been explored in the past, we would like to study how an optimization procedure on this manifold will compare to those approaches.

As demonstrated in \cite{wang2018identifying}, properties of the Hessian at the minima are also related to generalization of deep networks. Our framework provides a principled path to estimate properties of the Hessian such that they are invariant to rescaling of deep networks.

Finally, our framework can also be extended to nodewise rescalings of neural network parameters, as defined in \cite{neyshabur2015path}. 
For example, consider a neural network with 2 hidden layers with parameters $W=(W_1, W_2, W_3)$, represented as the function
$F_W(x) = W_3 \phi(W_2 \phi(W_1 x))$. 
For positive definite diagonal matrices  $\Lambda_1,\Lambda_2,\Lambda_3,\Lambda_4$ satisfying $\Lambda_1 \Lambda_2 = \Lambda_3 \Lambda_4 = I$, 
the network with parameters $T_\Lambda (W) = (\Lambda_1 W_1, \Lambda_3 W_2 \Lambda_2, \Lambda_4 W_3)$ implements the same function as the network with parameters $W$. For the nodewise rescalings of this nature, one can work with the following invariant metric on the parameter space:
\[ g(\eta, \xi) = \sum_{l=1}^{L} \sum_{i=1}^{d_l} \frac{\eta^i_l \xi^i_l}{(\textrm{vec}(W_l)^i)^2} \]
Using this metric in our framework will yield a flatness measure that is invariant to nodewise rescaling as well. 

\vspace{4mm}
\noindent {\bf Acknowledgements.~} We thank Rui Wu and Daniel Park for providing helpful comments on the paper. 

\appendix
\section{Missing Proofs from Section 2}
We retain the same notation from the main paper.

\begin{prop}
Let $W = (W_1, \ldots, W_L) \in \mathbb{R}^{d_1}_{*} \times \ldots \times \mathbb{R}^{d_L}_{*}$ be the parameters of a neural network with $L$ layers, and $\lambda = (\lambda_1, \ldots, \lambda_L) \in \mathbb{R}^{L}_{+}$ be a set of multipliers. Here $\mathbb{R}^{d_i}_{*} = \mathbb{R}^{d_i} \backslash \{0\} $. We can transform the layer weights by $\lambda$ in the following manner: $T_{\lambda}(W) = (\lambda_1 W_1, \ldots, \lambda_L W_L)$.
We introduce a relation from $\mathbb{R}^{n_1 \times n_0}_{*} \times \ldots \times \mathbb{R}^{n_L \times n_{L-1}}_{*}$ to itself, $W \sim Y$ if $\exists \lambda$ such that $Y = T_{\lambda} (W)$ and $\prod_{i=1}^L \lambda_i = 1$. The relation $\sim$, is an equivalence relation.
\end{prop}
\begin{proof}
\begin{enumerate}
\item It is self evident that $W \sim W$, with $\lambda = (1,\ldots, 1)$
\item If $W \sim Y$, then $\exists \lambda$ such that $Y = T_\lambda (W)$. Set $\tilde{\lambda} = (\lambda_1^{-1}, \ldots, \lambda_L^{-1})$, then $\tilde{\lambda}_i >0$ and $\prod_{i=1}^L \tilde{\lambda}_i = \frac{1}{\prod_{i=1}^L \lambda_i} = 1$. Also, $W = T_{\tilde{\lambda}} (Y)$, which means $Y \sim W$.
\item Let $W \sim Y$, and $Y \sim Z$. This means, $\exists \lambda^1$ such that $Y = T_{\lambda^1}(W)$, and $\exists \lambda^2$ such that $Z = T_{\lambda^2}(Y)$
Let $\tilde{\lambda} = (\lambda^1_1 \lambda^2_1, \ldots, \lambda^1_L \lambda^2_L)$. We see that $\tilde{\lambda}_i > 0$, and $\prod_{i=1}^L \tilde{\lambda}_i = \prod_{i=1}^L \lambda^1_i \times \prod_{j=1}^L \lambda^2_j = 1$. Since $Z = T_{\tilde{\lambda}} (W)$, we have that $W \sim Z$.
\end{enumerate}
Hence $\sim$ is an equivalence relation.
\end{proof}

\begin{prop}
The set $\mathcal{M}:= \overline{\mathcal{M}}/\sim$ obtained by mapping all points within an equivalence class to a single point in the set has a quotient manifold structure, making $\calM$ a differentiable quotient manifold.
\end{prop}
\begin{proof}
In order to prove that $\calM$ is a manifold, we need to show that:
\begin{enumerate}
\item $\textrm{graph}(\sim) = \{ (W,Y) : W,Y \in \overline{\mathcal{M}}, W \sim Y \}$ is an embedded submanifold of $\overline{\calM}\times \overline{\calM}$. \label{proof:manifold:point1}
\item The projection $\pi_1 : \textrm{graph}(\sim) \rightarrow \overline{\calM}$, $\pi_1(W,Y) = W$ is a submersion. \label{proof:manifold:point2}
\item $\textrm{graph}(\sim)$ is a closed subset of $\overline{\calM} \times \overline{\calM}$. \label{proof:manifold:point3}
\end{enumerate}
~\\
First, we look at a point $(W^0, Y^0) \in \textrm{graph}(\sim)$. This means $\exists \lambda \in \mathbb{R}^L_{+}$, $\prod_{i=1}^L \lambda_i = 1$, such that $Y^0 = T_{\lambda}(W^0)$. For every $V \in \mathbb{R}^{d_1} \times \ldots \times \mathbb{R}^{d_L}$ we can define $\gamma (t) = (W^0 + tV, T_\lambda ( W^0 + tV) )$ which is a smooth curve and an injection from $\mathbb{R}$ to $\textrm{graph}(\sim)$, and $\pi_1 (\gamma (t)) = W^0 + tV$. Since $\frac{d \pi_1 (\gamma (t)}{dt} = V$, we see that $\textrm{dim}(\textrm{range}(D\pi_1)) = \textrm{dim}(\overline{\calM})$, where $D\pi_1$ is the Jacobian of $\pi_1$. This means that $\pi_1$ is a submersion, proving point \ref{proof:manifold:point2}.

Next we will prove point \ref{proof:manifold:point3}. For this, we define a function $F : \overline{\calM} \times \overline{\calM} \rightarrow \mathbb{R}^{d_1 } \times \ldots \times \mathbb{R}^{d_L} \times \mathbb{R}$

\[ 
F(W,Y) = \begin{bmatrix}
Y_1 - \frac{\langle W_1, Y_1 \rangle}{||\textrm{vec}(W_1)||_2^2}W_1 \\
\vdots \\ 
Y_L - \frac{\langle W_L, Y_L \rangle}{||\textrm{vec}(W_L)||_2^2}W_L \\
\textrm{log} \left(\frac{\prod_{i=1}^L ||\textrm{vec}(Y_i)||_2^2}{\prod_{j=1}^L ||\textrm{vec}(W_j)||_2^2} \right)
\end{bmatrix}
\]

Under $F$, the preimage of $0_{d_1 \times \ldots \times d_L \times 1}$, is $\textrm{graph}(\sim)$. Since the preimage of a closed set is a closed set, we have that $\textrm{graph}(\sim)$ is a closed subset of $\overline{\calM} \times \overline{\calM}$.

Finally we will prove \ref{proof:manifold:point1}, by defining a submersion from $\overline{\calM}$ to $\mathbb{R}^{d_1-1 \times \ldots \times d_L-1 \times 1}$. Suppose there is a smooth function $F_1$ from $\overline{\calM}$ to $\textrm{St}(d_1-1, d_1) \times \ldots \times \textrm{St}(d_L-1, d_L)$ (where $\textrm{St}(p,n)$ is the $p$-dimensional Stiefel manifold), such that:
\[ F_1 (W) =  \begin{bmatrix}
W_1^{\perp} \\
\vdots \\
W_L^{\perp}
\end{bmatrix} \]
Here $W_i^\perp$ is an orthogonal basis for the $d_i - 1$ dimensional subspace that is orthogonal to $\textrm{vec}(W_i)$, for all $W \in \overline{\calM}$. Such an $F_1$ always exists, since given $W_i$ we can find $W_i^\perp$ by performing a Gram-Schmidt orthogonalization on $[\textrm{vec}(W_i) | E]$ and taking the last $d_i-1$ columns. Here $E$ is chosen such that $[\textrm{vec}(W_i) | E]$ is full rank. 

Now given $F_1$, we can define $F_2 : \overline{\calM} \times \overline{\calM} \rightarrow \mathbb{R}^{d_1 - 1} \times \ldots \times \mathbb{R}^{d_L - 1} \times \mathbb{R}$
\begin{equation*}
\begin{split}
F_2 (W,Y) = & \begin{bmatrix}
(W_1^\perp)^\top  \textrm{vec} (Y_1)\\ 
	\vdots \\
(W_L^\perp)^\top  \textrm{vec} (Y_L) \\
\sum_{i=1}^L \textrm{log} \frac{\langle \textrm{vec}(W_i), \textrm{vec}(Y_i) \rangle}{||\textrm{vec}(W_i)||_2^2}
\end{bmatrix}
\end{split}
\end{equation*}
For any $[X_1, \ldots, X_L,x] \in \mathbb{R}^{d_1 - 1} \times \ldots \times \mathbb{R}^{d_L - 1} \times \mathbb{R}$, we can define $\tilde{Y}$ such that 

\[ \textrm{vec}(\tilde{Y}) = \left[ W_1^\perp X_1 + \frac{x}{L}\textrm{vec}(Y_1), \ldots, W_L^\perp X_L + \frac{x}{L}\textrm{vec}(Y_L) \right] \]

which means, for points $(W,Y) \in \textrm{graph}(\sim)$:
\begin{equation*}
\begin{split}
DF_2 (W,Y) [0, \tilde{Y} ] &= \Bigg[ (W_1^\perp)^\top W_1^\perp X_1 + \frac{x}{L} (W_1^\perp)^\top \textrm{vec}(Y_1), \\
&\ldots, (W_L^\perp)^\top W_L^\perp X_L + \frac{x}{L} (W_L^\perp)^\top \textrm{vec}(Y_L), \\
&\sum_{i=1}^L \frac{||\textrm{vec}(W_i)||_2^2}{ \langle \textrm{vec}(W_i), \textrm{vec}(Y_i)\rangle} \\
&\times \frac{\textrm{vec}(W_i)^\top \left( W_i^\perp X_i + \frac{x}{L}\textrm{vec}(Y_i)\right)}{||\textrm{vec}(W_i)||_2^2} \Bigg] \\
&= [X_1, \ldots, X_L, x]
\end{split}
\end{equation*}
This means that $F_2$ is a submersion at each point of $\textrm{graph}(\sim)$, and the set $F_2^{-1}(0) = \textrm{graph}(\sim)$ is an embedded submanifold of $\overline{\calM} \times \overline{\calM}$. This concludes our proof that $\calM = \overline{\calM}/\sim$ is a quotient manifold.
\end{proof}

\subsection{Deep Networks with Biases}
We recall that Deep Networks with biases are defined as follows:

\begin{equation*}
\begin{split}
F_{(W,b)}(\mathrm{x}) = &W_L \phi_{L-1} ( W_{L-1} \phi_{L-2}(W_{L-2}\ldots\phi_{1}(W_1 \mathrm{x} + b_1) \\ &\ldots +b_{L-2})+b_{L-1} ) +b_L
\end{split}
\end{equation*}

The equivalence relation for the parameter space of deep networks with biases is defined through the following transformation. 
Suppose we have $\lambda_i \in \mathbb{R}_{+}, i=1, \ldots, L$, such that $\prod_{i=1}^L \lambda_i = 1$. Consider the following transformation:

\begin{equation*}
\begin{split}
T_{\lambda} ((W,b)) &= (\lambda_L W_L, \ldots, \lambda_1 W_1, \\ &\prod_{i=1}^L \lambda_i b_L, \prod_{i=1}^{L-1} \lambda_i b_{L-1}, \ldots, \lambda_1 b_1)
\end{split}    
\end{equation*}

Now, if $(Y,c) = T_{\lambda}((W,b))$, then $F_{(W,b)} ( \mathrm{x}) = F_{(Y,c)} ( \mathrm{x}), \forall \mathrm{x}\in \mathbb{R}^{n_0}$. Thus we define the equivalence relation $\sim$, where $(Y,c) \sim (W,b)$ if $\exists \lambda$ such that $(Y,c) = T_{\lambda}((W,b))$. 

Let us denote $\overline{\calM}_i = \mathbb{R}^{d_i} \times \mathbb{R}^{n_i}$, as the Euclidean space for each layer. The product space $\overline{\calM} = \overline{\calM}_1 \times \ldots \times \overline{\calM}_L$ is the entire space of parameters for neural networks with biases.

\begin{prop}
The set $\mathcal{M}:= \overline{\mathcal{M}}/\sim$ obtained by mapping all points within an equivalence class to a single point in the set has a quotient manifold structure, making $\calM$ a differentiable quotient manifold.
\end{prop}
\begin{proof}
In order to prove that $\calM$ is a manifold, we need to show that:
\begin{enumerate}
\item $\textrm{graph}(\sim) = \{ ((W,b),(Y,c)) : (W,b),(Y,c) \in \overline{\mathcal{M}}, (W,b) \sim (Y,c) \}$ is an embedded submanifold of $\overline{\calM}\times \overline{\calM}$. \label{proof:bias:manifold:point1}
\item The projection $\pi_1 : \textrm{graph}(\sim) \rightarrow \overline{\calM}$, $\pi_1((W,b),(Y,c)) = (W,b)$ is a submersion. \label{proof:bias:manifold:point2}
\item $\textrm{graph}(\sim)$ is a closed subset of $\overline{\calM} \times \overline{\calM}$. \label{proof:bias:manifold:point3}
\end{enumerate}
~\\
First, we look at a point $((W^0,b^0),(Y^0,c^0)) \in \textrm{graph}(\sim)$. This means $\exists \lambda \in \mathbb{R}^L_{+}$, $\prod_{i=1}^L \lambda_i = 1$, such that $(Y^0,c^0) = T_{\lambda}((W^0,b^0))$. For every $(V,v) \in \mathbb{R}^{d_1} \times \ldots \times \mathbb{R}^{d_L} \times \mathbb{R}^{n_1}\times \ldots \times \mathbb{R}^{n_L}$ we can define $\gamma (t) = ((W^0,b^0) + t(V,v), T_\lambda ( (W^0,b^0) + t(V,v)) )$ which is a smooth curve and an injection from $\mathbb{R}$ to $\textrm{graph}(\sim)$, and $\pi_1 (\gamma (t)) = (W^0,b^0) + t(V,v)$. Since $\frac{d \pi_1 (\gamma (t)}{dt} = (V,v)$, we see that $\textrm{dim}(\textrm{range}(D\pi_1)) = \textrm{dim}(\overline{\calM})$, where $D\pi_1$ is the Jacobian of $\pi_1$. This means that $\pi_1$ is a submersion, proving point \ref{proof:bias:manifold:point2}.

Next we will prove point \ref{proof:bias:manifold:point3}. For this, we define a function $F : \overline{\calM} \times \overline{\calM} \rightarrow \mathbb{R}^{d_1 } \times \ldots \times \mathbb{R}^{d_L} \times \mathbb{R}^{n_1} \times \ldots \times \mathbb{R}^{n_L} \times \mathbb{R}^{L+1}$

\[ 
F((W,b),(Y,c)) = \begin{bmatrix}
Y_1 - \frac{\langle W_1, Y_1 \rangle}{||\textrm{vec}(W_1)||_2^2}W_1 \\
\vdots \\ 
Y_L - \frac{\langle W_L, Y_L \rangle}{||\textrm{vec}(W_L)||_2^2}W_L \\
c_1 - \frac{\langle b_1, c_1 \rangle}{||b_1||_2^2}b_1 \\
\vdots \\ 
c_L - \frac{\langle b_L, c_L \rangle}{||b_L||_2^2}b_L \\
\textrm{log} \left(\frac{\prod_{i=1}^L ||\textrm{vec}(Y_i)||_2^2}{\prod_{j=1}^L ||\textrm{vec}(W_j)||_2^2} \right) \\
\textrm{log} \left(\frac{||c_1||_2^2}{||b_1||_2^2} \right) - \textrm{log} \left(\frac{||\textrm{vec}(Y_1)||_2^2}{||\textrm{vec}(W_1)||_2^2} \right) \\
\textrm{log} \left(\frac{||c_2||_2^2}{||b_2||_2^2} \right) - \textrm{log} \left(\frac{||c_1||_2^2}{||b_1||_2^2} \times \frac{||\textrm{vec}(Y_2)||_2^2}{||\textrm{vec}(W_2)||_2^2} \right) \\
\vdots \\
\textrm{log} \left(\frac{||c_L||_2^2}{||b_L||_2^2} \right) - \textrm{log} \left(\frac{||c_{L-1}||_2^2}{||b_{L-1}||_2^2} \times \frac{||\textrm{vec}(Y_L)||_2^2}{||\textrm{vec}(W_L)||_2^2} \right)
\end{bmatrix}
\]

Under $F$, the preimage of $0_{d_1 \times \ldots \times d_L \times \times n_1 \times \ldots \times n_L \times L+1}$, is $\textrm{graph}(\sim)$. Since the preimage of a closed set is a closed set, we have that $\textrm{graph}(\sim)$ is a closed subset of $\overline{\calM} \times \overline{\calM}$.

Finally we will prove \ref{proof:bias:manifold:point1}, by defining a submersion from $\overline{\calM}$ to $\mathbb{R}^{d_1-1 \times \ldots \times d_L-1 \times n_1-1 \times \ldots \times n_L-1 \times L+1}$. Suppose there is a smooth function $F_1$ from $\overline{\calM}$ to $\textrm{St}(d_1-1, d_1) \times \ldots \times \textrm{St}(d_L-1, d_L) \times \textrm{St}(n_1-1, n_1) \times \ldots \times \textrm{St}(n_L-1, n_L)$ (where $\textrm{St}(p,n)$ is the $p$-dimensional Stiefel manifold), such that:
\[ F_1 ((W,b)) =  \begin{bmatrix}
W_1^{\perp} \\
\vdots \\
W_L^{\perp} \\
b_1^{\perp} \\
\vdots \\
b_L^{\perp}
\end{bmatrix} \]
Here $W_i^\perp$ is an orthogonal basis for the $d_i - 1$ dimensional subspace that is orthogonal to $\textrm{vec}(W_i)$, and $b_i^\perp$ is an orthogonal basis for the $n_i - 1$ dimensional subspace orthogonal to $b_i$, for all $(W,b) \in \overline{\calM}$. Such an $F_1$ always exists, since given $W_i$ (alternatively $b_i$) we can find $W_i^\perp$ (alternatively $b_i^\perp$) by performing a Gram-Schmidt orthogonalization on $[\textrm{vec}(W_i) | E]$ (or $[b_i | E]$) and taking the last $d_i-1$ (or $n_i-1$) columns. Here $E$ is chosen such that $[\textrm{vec}(W_i) | E]$ (or $[b_i | E]$) is full rank. 

Now given $F_1$, we can define $F_2 : \overline{\calM} \times \overline{\calM} \rightarrow \mathbb{R}^{d_1 - 1} \times \ldots \times \mathbb{R}^{d_L - 1} \times \mathbb{R}$
\begin{equation*}
\begin{split}
F_2 ((W,b),(Y,c)) = & \begin{bmatrix}
(W_1^\perp)^\top  \textrm{vec} (Y_1)\\ 
	\vdots \\
(W_L^\perp)^\top  \textrm{vec} (Y_L) \\
(b_1^\perp)^\top  c_1\\ 
	\vdots \\
(b_L^\perp)^\top  c_L \\
\sum_{i=1}^L \textrm{log} \frac{\langle \textrm{vec}(W_i), \textrm{vec}(Y_i) \rangle}{||\textrm{vec}(W_i)||_2^2} \\
\textrm{log} \left(\frac{\langle b_1, c_1 \rangle}{||b_1||_2^2} \right) - \textrm{log} \left(\frac{\langle \textrm{vec}(W_1), \textrm{vec}(Y_1) \rangle}{||\textrm{vec}(W_1)||_2^2} \right) \\
\textrm{log} \left(\frac{\langle b_2, c_2 \rangle}{||b_2||_2^2} \right) - \textrm{log} \left(\frac{\langle b_1, c_1 \rangle}{||b_1||_2^2} \times \frac{\langle \textrm{vec}(W_1), \textrm{vec}(Y_1) \rangle}{||\textrm{vec}(W_1)||_2^2} \right) \\
\vdots \\
\textrm{log} \left(\frac{\langle b_L, c_L \rangle}{||b_L||_2^2} \right) - \textrm{log} \left(\frac{\langle b_{L-1}, c_{L-1} \rangle}{||b_{L-1}||_2^2} \times \frac{\langle \textrm{vec}(W_{L-1}), \textrm{vec}(Y_{L-1}) \rangle}{||\textrm{vec}(W_{L-1})||_2^2} \right)\end{bmatrix}
\end{split}
\end{equation*}
For any $[X_1, \ldots, X_L, z_1, \ldots, z_L, x, u_1, \ldots u_L] \in \mathbb{R}^{d_1 - 1} \times \ldots \times \mathbb{R}^{d_L - 1} \times \mathbb{R}^{n_1 - 1} \times \ldots \times \mathbb{R}^{n_L - 1} \times \mathbb{R}^{L+1}$, we can define $(\tilde{Y}, \tilde{c})$ such that 

\[ \textrm{vec}(\tilde{Y}) = \left[ W_1^\perp X_1 + \frac{x}{L}\textrm{vec}(Y_1), \ldots, W_L^\perp X_L + \frac{x}{L}\textrm{vec}(Y_L) \right] \]

\begin{equation*}
\begin{split}
\tilde{c} =& \Bigg[ b_1^\perp z_1 + \left(u_1 + \frac{x}{L} \right) c_1, \\
& b_2^\perp z_2 + \left(u_2 + u_1 + \frac{2x}{L} \right) c_2, \\
& \ldots, \\
& b_L^\perp z_L + \left(u_L + u_{L-1}+ \frac{Lx}{L} \right) c_L \Bigg]
\end{split}
\end{equation*}

which means, for points $(W,Y) \in \textrm{graph}(\sim)$:
\begin{equation*}
\begin{split}
DF_2 ((W,b),(Y,c)) [0, (\tilde{Y}, \tilde{c}) ] =& [X_1, \ldots, X_L, z_1, \ldots, z_L,\\
& x, u_1, \ldots, u_L]
\end{split}
\end{equation*}
This means that $F_2$ is a submersion at each point of $\textrm{graph}(\sim)$, and the set $F_2^{-1}(0) = \textrm{graph}(\sim)$ is an embedded submanifold of $\overline{\calM} \times \overline{\calM}$. This concludes our proof that $\calM = \overline{\calM}/\sim$ is a quotient manifold.
\end{proof}

\bibliography{references}
\bibliographystyle{abbrv}

\end{document}